\documentclass[10pt, a4paper]{article}

\usepackage{lrec-coling2024} % this is the new style
\usepackage{booktabs}
\usepackage{multirow}
\usepackage{adjustbox}
\usepackage{rotating}
\usepackage{scalerel} 
\title{\raisebox{-.25\height}{\includegraphics[width=0.7cm]{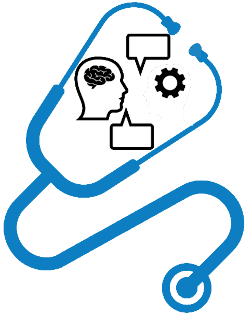}} Medical mT5: An Open-Source Multilingual Text-to-Text LLM \\ for The Medical Domain}

\name{Iker García-Ferrero$^{1}$, Rodrigo Agerri$^{1}$, Aitziber Atutxa$^{1}$, Elena Cabrio$^{2}$, \\
{\bf \large Iker de la Iglesia$^{1}$, Alberto Lavelli$^{3}$, Bernardo Magnini$^{3}$, Benjamin Molinet$^{2}$,  } \\ 
{\bf \large Johana Ramirez-Romero$^{4}$, German Rigau$^{1}$, Jose Maria Villa-Gonzalez$^{4}$,}\\
{\bf \large Serena Villata$^{2}$, Andrea Zaninello$^{3}$}}

\address{$^{1}$HiTZ Center - Ixa, University of the Basque Country UPV/EHU \\ 
$^{2}$Université Côte d’Azur, CNRS, Inria, I3S, France\\
$^{3}$Fondazione Bruno Kessler, Via Sommarive 18, Povo, Trento, Italy \\
$^{4}$Cruces University Hospital, (Barakaldo, Biscay, Spain) \\
         \{iker.garciaf, rodrigo.agerri\}@ehu.eus,\{FirstName.LastName\}@univ-cotedazur.fr, \{lastName\}@fbk.eu\\ }

\abstract{Research on language technology for the development of medical 
	applications is currently a hot topic in Natural Language Understanding and
			 Generation. Thus, a number of large language models (LLMs) 
			 have recently been adapted to the medical domain, so that they can be
			 used as a tool for mediating in human-AI interaction. While these LLMs display
			 competitive performance on automated medical texts benchmarks,
			 they have been pre-trained and evaluated with a focus on a single language (English
			 mostly). This is particularly true of text-to-text models, which
			 typically require large amounts of domain-specific pre-training
			 data, often not easily
			 accessible for many languages. In this paper, we address these shortcomings by
			 compiling, to the best of our knowledge, the largest multilingual
			 corpus for the medical domain in four languages,
			 namely English, French, Italian and Spanish. This new corpus has been
			 used to train Medical mT5, the first open-source text-to-text
			 multilingual model for the medical domain. Additionally, we
			 present two new evaluation benchmarks for all four languages with
			 the aim of facilitating multilingual research in this domain.
			 A comprehensive evaluation shows that Medical mT5 outperforms both
			 encoders and similarly sized text-to-text models for the Spanish, French, and Italian 
			 benchmarks, while being competitive with current state-of-the-art
			 LLMs in English.\\ 
		 \newline  }

\begin{document}

\maketitleabstract

\section{Introduction}

As it is the case for many application domains, there is an increasing interest
in applying Artificial Intelligence (AI) and Natural Language Processing (NLP)
techniques to assist medical experts in their everyday activities.  With this
aim in mind, in the last few years a number of large language models (LLMs)
have been trained or adapted to the medical
domain. These include encoder models such as SciBERT \cite{beltagy2019scibert},
BioBERT \cite{DBLP:journals/bioinformatics/LeeYKKKSK20} or PubmedBERT
\cite{DBLP:journals/health/GuTCLULNGP22}. While these models have obtained
state-of-the-art results in discriminative tasks, they are typically smaller in scale
and scope with respect to medical text-to-text models such as SciFive
\cite{DBLP:journals/corr/abs-2106-03598}, BioGPT \cite{10.1093/bib/bbac409}
Med-PaLM \cite{singhal-palm}, PMC-LLaMA \cite{wu2023pmcllama} or ClinicalGPT
\cite{Wang2023ClinicalGPTLL}.

However, the development of all the aforementioned text-to-text LLMs
has been focused on a single language, usually English. As a
consequence, there is a lack of high-quality multilingual
evaluation benchmarks for the medical domain. Thus, although there
have been efforts to generate evaluation data in languages other than English
\cite{Wang2023ClinicalGPTLL,carrino-etal-2022-pretrained}, they have consisted
largely in monolingual approaches.

In order to address these issues, we have compiled, to the best of our
knowledge, the largest multilingual corpus for training
LLMs adapted to the medical domain. Our corpus includes 3B words in four languages, namely,
English, Spanish, French, and Italian. While relatively small when compared
to English existing datasets \cite{wu2023pmcllama}, it allowed us to build Medical mT5, the first open-source
text-to-text multilingual model for the medical domain. 

Medical mT5 is an encoder-decoder model developed by continuing the training of
publicly available mT5 \cite{xue-etal-2021-mt5} checkpoints on medical domain
data for English, Spanish, French, and Italian. Additionally, we have also created
two new multilingual sequence labeling (argument component detection) and generative question
answering datasets for the evaluation of multilingual LLMs in the medical
domain.

A comprehensive experimental evaluation shows that Medical mT5 outperforms
similarly-sized text-to-text models for the Spanish, French, and Italian
benchmarks while being competitive in English with respect to current
state-of-the-art text-to-text \cite{xue-etal-2021-mt5,chung-flan-instruction-models} and encoder-only models
\cite{DBLP:journals/bioinformatics/LeeYKKKSK20,debertav3}. The results show
that continuing pre-training of a multilingual text-to-text model such as mT5 allows to
successfully adapt it to the medical domain, even when the amount of
domain-specific data is relatively modest (ranging between 1B words for English
and Spanish to 150M in Italian). Summarizing, the contributions of our work are the following:
(i) the collection of the largest publicly available in-domain medical multilingual corpus 
	for Spanish, French, and Italian languages. Together with the already existing English data, 
	we release a corpus of 3 billion tokens. \footnote{\url{https://hf.co/datasets/HiTZ/Multilingual-Medical-Corpus}}
(ii) two new datasets for Spanish, French, and Italian on Argument Mining\footnote{\url{https://hf.co/datasets/HiTZ/multilingual-abstrct}} and
	generative Question Answering tasks, generated taking their original English versions as a
	starting point. \footnote{\url{https://hf.co/datasets/HiTZ/Multilingual-BioASQ-6B}}
(iii) the public release of two Medical mT5 versions: a 770M \footnote{\url{https://hf.co/HiTZ/Medical-mT5-large}}  and 3B \footnote{\url{https://hf.co/HiTZ/Medical-mT5-xl}} parameter text-to-text open-source models which obtain state-of-the-art results in multilingual sequence labelling for the medical domain, most notably in multi-task and zero-shot crosslingual settings. 

Other benefits of our Medical mT5 models include the comparatively low hardware
requirements needed for both fine-tuning on downstream tasks (the large 770M version easily fits in a 24GB V100 GPU) and for inference (a 12GB GPU should be enough). As an example, a LLaMA 7B model \cite{wu2023pmcllama} requires at least a 80GB A100 GPU using LoRA \cite{hu2021lora} or a more demanding 4 80GB A100 GPUs without it. Code, data, models, and benchmarks are publicly available to facilitate reproducibility of results and encourage future multilingual research on the medical domain.

\section{Related Work}

As it has been the case in most application domains, Large Language Models (LLMs) have facilitated huge improvements
in the state-of-the-art for medical NLP tasks \cite{singhal-palm,wu2023pmcllama,mayer2021enhancing}. The most popular approaches are those that use models pre-trained on medical corpora such as SciBERT \cite{beltagy2019scibert}, BioBERT \cite{DBLP:journals/bioinformatics/LeeYKKKSK20}, PubmedBERT \cite{DBLP:journals/health/GuTCLULNGP22}, BSC-BIO \cite{carrino-etal-2022-pretrained} or BioLinkBERT \cite{DBLP:conf/acl/YasunagaLL22}.

While the previous encoder-only models focused on discriminative tasks, the emergence of generative models such as LLaMa \cite{touvron2023llama}, PaLM \cite{singhal-palm} or GPT-3 \cite{brown2020language} has resulted in a huge interest in adapting such LLMs to the medical domain. These models, to name but a few, include SciFive \cite{DBLP:journals/corr/abs-2106-03598}, and English T5 encoder-decoder model adapted to the scientific domain, and decoder models such as BioGPT \cite{10.1093/bib/bbac409}, Med-PaLM \cite{singhal-palm}, PMC-LLaMA \cite{wu2023pmcllama} and ClinicalGPT \cite{Wang2023ClinicalGPTLL}. 

Additionally, a range of Abstractive Question Answering tasks have been proposed as evaluation benchmarks on which the larger models \cite{wu2023pmcllama,singhal-palm,Wang2023ClinicalGPTLL} obtain best results. While interesting, both these LLMs and benchmarks have been developed with a focus on a single language, usually English. 
Furthermore, these LLMs require hardware which is simply not affordable for the large majority of end-users and researchers. In order to address these issues, we propose Medical mT5, a multilingual text-to-text model adapted to the medical domain which, despite its relatively modest size and cheap running requirements, obtains competitive results, most notably in multi-task and zero-shot cross-lingual settings.

\section{Compiling a Multilingual Corpus for the Medical Domain}\label{sec:corpus}

Obtaining good quality medical corpora is usually difficult due to the sensitive nature of the data. This is even more challenging for non-English languages, as the availability of data for other languages is in general more restricted. Despite these issues, we have successfully gathered and curated a diverse collection of public relevant corpora of medical texts in English, French, Italian and Spanish to generate the Medical mT5 model.

\subsection{English}

As listed in Table \ref{tab:english-data}, we collected around 1B words from three sources related to the medical domain: (i) \textbf{ClinicalTrials} is a set of documents of clinical studies from all over the world \citeplanguageresource{clinical_trials}; (ii) \textbf{EMEA} is an English-Spanish parallel corpus with documents provided by the European Medicines Agency \cite{TIEDEMANN12.463} and, (iii) PubMed \citeplanguageresource{pubmed}, which contains data from various sources such as MEDLINE, life science journals and online books, provides the bulk of the English data.

\begin{table}[ht]
\centering
\begin{tabular}{l|r}
\toprule
\textbf{Source} & \textbf{Words} \\
\midrule
ClinicalTrials & 127.4M \\
EMEA & 12M \\
PubMed & 968.4M \\
\bottomrule
\end{tabular}
\caption{English data sources and word counts.}
\label{tab:english-data}
\end{table}

\subsection{Spanish}

Apart from \textbf{EMEA} and \textbf{PubMed}, which we also used for Spanish, the biggest portion of the data came from the \textbf{Medical Crawler}, a biomedical corpus compiled by \citet{carrino-etal-2022-pretrained}. Additionally, we also included \textbf{SPACC} \citeplanguageresource{spacc}, \textbf{UFAL} \citeplanguageresource{ufal_medical_corpus} and \textbf{WikiMed}, a corpus built ad-hoc from Wikipedia entries. Table \ref{tab:spanish-data} provides the details of the collected data, which amounts to $\approx$1B words.

\begin{table}[ht]
\centering
\begin{tabular}{l|r}
\toprule
\textbf{Source} & \textbf{Words} \\
\midrule
EMEA & 13.6M \\
PubMed & 8.4M \\
Medical Crawler & 918M \\
SPACC & 350K \\
UFAL & 10.5M \\
WikiMed & 5.2M \\
\bottomrule
\end{tabular}
\caption{Spanish data sources and word counts.}
\label{tab:spanish-data}
\end{table}

\subsection{French}
%In order to make our model able to interact with the French language in the medical domain, it was essential to gather a sizable volume of validated data.
A total of 7,192,779 sentences and 670,972,717 words were compiled using the data sources listed in Table~\ref{tab:dataSources}.

\begin{table}[ht]
\centering
\begin{tabular}{l|r}
\toprule
\textbf{Source} & \textbf{Words} \\
\midrule
PubMed & 1.4M \\
Science Direct & 15.2M \\
Wikipedia - Médecine & 5M \\
EDP & 48K \\
Google Patents & 654M \\
\bottomrule
\end{tabular}
\caption{French data sources and word counts.}
\label{tab:dataSources}
\end{table}

\textbf{PubMed} data was extracted using the \texttt{Bio.Entrez} package\footnote{\url{https://biopython.org/docs/1.75/api/Bio.Entrez.html}}. \textbf{Science Direct} offers a collection of scientific and medical publications which can be extracted via their the official API\footnote{\url{https://dev.elsevier.com/}}. We filtered relevant articles with the keyword ``Médecine'', and the obtained XML documents were parsed to extract the \texttt{<dc:description>} tag.
As for Spanish, we took advantage of \textbf{Wikipedia} as a source of medical knowledge to obtain HTML formatted data from the category ``Category:Médecine''. The \textbf{EDP French/English Parallel Medical Corpus}~\cite{DBLP:conf/wmt/Jimeno-YepesNNV17} provides bilingual content from journals that address domains such as dentistry and life sciences. From this source, we downloaded the dataset labeled ``EDP French corpus, text format''. Finally, \textbf{Google Patents} is a comprehensive repository of patent data from around the world. Google Patents data were retrieved by filtering using the IPC code and abstract language. 
%The data request was queried on Google Big Query (details on the query in Appendix upon publication). 
%with the following request:

%{\small
%\begin{verbatim}
%SELECT publication_number, abstract.text
%FROM `patents-public-data.patents.publications`, 
%  UNNEST(abstract_localized) as abstract, 
%  UNNEST(ipc) as ipc
%WHERE abstract.language = "fr" 
%  AND (ipc.code LIKE 'A61B%' 
%    OR ipc.code LIKE 'A61C%'
%    OR ipc.code LIKE 'A61F%'
%    OR ipc.code LIKE 'A61H%'
%    OR ipc.code LIKE 'A61K%'
%    OR ipc.code LIKE 'A61L%'
%    OR ipc.code LIKE 'A61M%'
%    OR ipc.code LIKE 'A61P%')
%\end{verbatim}}

%Every collected data was cleaned and converted into JSONL format. 
%. These transformations involved the use of regex patterns to eliminate HTML tags and rectify encoding anomalies. 
A final French language verification step was undertaken by applying the \texttt{langdetect} package (version 1.0.9).

\subsection{Italian}

The crawling and pre-processing of the Italian split of the corpus followed the methodology described by \citet{carrino-etal-2022-pretrained}. First, we compiled a list of 504 medical terms, which we use as seeds to scrape the Italian split of the \textbf{MC4 Common Crawl Corpus} \citeplanguageresource{common_crawl} by only selecting the pages which contained at least one of the keywords in their URL domain. To create the list, we extracted 600 keyword terms related to medicine from the \textit{Dizionario analogico della Lingua Italiana} (Zanichelli). We excluded some sectors and discarded terms that may lead to ambiguous queries (e.g., actions, which contained mainly verbs, proverbs, general terms like ``assistente'', etc.). We normalized rare variants (``bacteriologia'' to ``batteriologia'') and stemmed all terms without lemmatizing, as most terms are already lemmatized in the dictionary; we performed univerbation of multiword units (e.g., ``esamedelleurine'', ``follow-up''), and removed the duplicates.
%which included removal of spurious tags, references to figures and tables, etc.,
This resulted in a corpus of 67 million tokens, which we joined with other sources of text such as \textbf{Medical dissertations}, \textbf{Drug use instructions}, \textbf{PubMed abstracts}, etc. as detailed in Table \ref{tab:dataSources_it}, resulting in a $\approx$145M word corpus.

\begin{table}[ht]
\centering
\begin{tabular}{l|r}
\toprule
\textbf{Source} & \textbf{Words} \\
\midrule
Medical Commoncrawl - IT & 67M \\
Drug instructions & 30.5M \\
Wikipedia - Medicina & 13.3M \\
E3C Corpus - IT & 11.6M \\
Medicine descriptions & 6.3M \\
Medical theses & 5.8M \\
Medical websites & 4M \\
PubMed & 2.3M \\
Supplement description & 1.3M \\
Medical notes & 975K \\
Pathologies & 157K \\
Medical test simulations & 26K \\
Clinical cases & 20K \\
\bottomrule
\end{tabular}
\caption{Italian data sources and word counts.}
\label{tab:dataSources_it}
\end{table}
\clearpage

\section{\includegraphics[width=0.7cm]{logo-antidote.png} Medical mT5}

%In this section, we describe our approach to develop the Medical mT5 model. Medical mT5 is based on MT5 model \cite{xue-etal-2021-mt5}.
%\subsection{mT5 models}

Multilingual T5 (mT5) \cite{xue-etal-2021-mt5} is an extension of the original T5 \cite{DBLP:journals/jmlr/RaffelSRLNMZLL20} framework, which is optimized for multilingual tasks. The T5 model is grounded in the transformer encoder-decoder architecture \cite{DBLP:conf/nips/VaswaniSPUJGKP17}. With its decoder block, T5 is capable of generating sequences of tokens in an auto-regressive fashion. T5 was designed to convert every NLP problem into a text-to-text task, and mT5 extends this strategy to a multitude of languages, leveraging a shared vocabulary for diverse scripts. mT5 was trained using mC4, a 1 Trillion token Common Crawl-based dataset covering 101 languages. The pre-training is based on a masked language modeling ``span-corruption'' objective, where consecutive spans of input tokens are replaced with a mask and the model is trained to reconstruct the masked-out tokens.

\subsection{Pre-training Medical mT5}

Medical mT5 is built upon the same architecture as mT5 \cite{xue-etal-2021-mt5}. We release two diffent models: Medical-mT5-large (738M parameters) and Medical-mT5-xl (3 billion parameters). Both models were initialized using the pre-trained weights of their corresponding mT5 checkpoints and continued their pre-training using the 3B word medical domain dataset described in Section \ref{sec:corpus} (with x2 oversampling for the Italian split). To prevent over-fitting, we run the training for only one epoch, as preliminary experiments showed that performance degraded with more epochs. We adhered to the self-supervised parameter settings recommended by \citet{xue-etal-2021-mt5} and detailed in Table \ref{tab:PreTraining}. It should be noted that Medical-mT5-large was trained with a sequence length of 1024 tokens whereas Medical-mT5-xl was limited to a sequence length of 480 tokens due to GPU memory limitations. Medical mT5 was trained using the Flax implementation of mT5 in the Hugging Face transformers library \cite{wolf-etal-2020-transformers}. All experiments were conducted on our private servers, employing 4xA100 80GB GPUs. We made calculations for a carbon footprint estimation based on a 400W consumption per GPU and a carbon intensity of 0.171 kg/kWh\footnote{Sourced from \url{https://app.electricitymaps.com/map}}. 

\begin{table}[htb]
\centering

\adjustbox{max width=\linewidth}{
\begin{tabular}{@{}lrr@{}}
\toprule
 & Medical-mT5-large & Medical-mT5-xl \\ \midrule
Param. no. & 738M & 3B \\
Sequence Lenght & 1024 & 480 \\
Token/step & 65536 & 30720 \\
Epochs & 1 & 1 \\
Total Tokens & 4.5B & 4.5B \\
Optimizer & Adafactor & Adafactor \\
LR & 0.001 & 0.001 \\
Scheduler & Constant & Constant \\
Hardware & 4xA100 & 4xA100 \\
Time (h) & 10.5 & 20.5 \\
CO\textsubscript{2}eq (kg) & 2.9 & 5.6 \\ \bottomrule
\end{tabular}}
\caption{Pre-Training settings for Medical mT5.}
\label{tab:PreTraining}
\end{table}

\section{Generating New Multilingual Benchmarks}\label{sec:new-benchmarks}

The lack of multilingual evaluation benchmarks for the medical domain motivated us to generate new evaluation data for our languages of interest, as only the relatively small E3C \cite{e3c} was already available for all 4 languages.  We focused on two different types of tasks: (i) a sequence labelling task, \textbf{Argument Mining}, consisting in detecting and classifying the argument component spans and their relations, (ii) \textbf{Abstractive Question Answering}, where the model is expected to generate an answer in response to an input question. In both cases we took existing English labelled data as a starting point.

\begin{table*}[htb]
\centering
\small
\adjustbox{max width=\textwidth}{
\begin{tabular}{@{}ccccc@{}}
\toprule
Representation & Task & Dataset & Languages & Entity Type \\ \midrule
 & \cellcolor{ForestGreen!10} & \cellcolor{ForestGreen!10}NCBI-Disease \cite{ncbi-disease} & \cellcolor{ForestGreen!10}EN & \cellcolor{ForestGreen!10}Disease \\
 & \cellcolor{ForestGreen!10} & \cellcolor{ForestGreen!10}BC5CDR Disease \cite{bc5cdr} & \cellcolor{ForestGreen!10}EN & \cellcolor{ForestGreen!10}Disease \\
 & \cellcolor{ForestGreen!10} & \cellcolor{ForestGreen!10}BC5CDR Chemical \cite{bc5cdr} & \cellcolor{ForestGreen!10}EN & \cellcolor{ForestGreen!10}Chemical \\
 & \cellcolor{ForestGreen!10} & \cellcolor{ForestGreen!10}DIANN \cite{diann} & \cellcolor{ForestGreen!10}EN, ES & \cellcolor{ForestGreen!10}Disability \\
 & \cellcolor{ForestGreen!10} & \cellcolor{ForestGreen!10}E3C \cite{e3c} & \cellcolor{ForestGreen!10}EN, ES, FR, IT & \cellcolor{ForestGreen!10}Clinical Entity \\
 & \multirow{-6}{*}{\cellcolor{ForestGreen!10}\begin{tabular}[c]{@{}c@{}}Named Entity \\ Recognition\end{tabular}} & \cellcolor{ForestGreen!10}PharmaCoNER \cite{pharmaconer} & \cellcolor{ForestGreen!10}ES & \cellcolor{ForestGreen!10}Pharmacological \\
\multirow{-7}{*}{\begin{tabular}[c]{@{}c@{}}Sequence\\ Labelling\end{tabular}} & \cellcolor{CornflowerBlue!10}\begin{tabular}[c]{@{}c@{}}Argument \\ Mining\end{tabular} & \cellcolor{CornflowerBlue!10}AbstRCT \cite{mayer2021enhancing} & \cellcolor{CornflowerBlue!10}EN, ES, FR, IT & \cellcolor{CornflowerBlue!10}Claims and Premises \\ \midrule
\begin{tabular}[c]{@{}c@{}}Generative \\ Question Answering\end{tabular} & \cellcolor{Yellow!10}\begin{tabular}[c]{@{}c@{}}Question \\ Answering\end{tabular} & \cellcolor{Yellow!10}BioASQ 6B \cite{bioasq} & \cellcolor{Yellow!10}EN, ES, FR, IT & \cellcolor{Yellow!10}Biomedical QA \\ \bottomrule
\end{tabular}}
\caption{List of evaluation tasks used to measure the performance of Medical mT5.}
\label{tab:tasks}

\end{table*}

\subsection{Argument Mining}

The AbstRCT dataset is composed by English medical and scientific texts collected from the MEDLINE database and manually annotated with two types of argument components: Claims and Premises \cite{mayer2021enhancing}. 

A `claim'  is a concluding statement made by the author about the outcome of the study. In the medical domain it may be an assertion of a diagnosis or a treatment. A `premise' corresponds to an observation or measurement in the study (ground truth), which supports or attacks another argument component, usually a claim. It is important that they are observed facts, therefore, credible without further evidence.

We generated French and Italian parallel versions of the dataset using the same method as for Spanish, based on machine translation and semi-manual annotation projection \cite{yegingbergenova-cross}. The AbstRCT dataset is divided in three splits, neoplasm, glaucoma and mixed. Following previous work, we fine-tune the models with the first one and then evaluate the in-domain performance on the neoplasm test split and the cross-domain performance on the glaucoma and mixed splits. Previous works using the AbstRCT datasets have employed different definitions of the $F_1$ score metric, such as token-level $F_1$ \cite{mayer2021enhancing,yegingbergenova-cross}. However, in this paper we report results using the standard sequence level $F_1$ score \cite{tjong-kim-sang-de-meulder-2003-introduction}, a much more strict metric, which explains the lower results for all the models.

\subsection{Question Answering}\label{sec:QA_explained}

We use the BioASQ-6B English Question Answering dataset \cite{bioasq} to generate parallel French, Italian and Spanish versions. Given a biomedical question and a set of snippets of text with relevant information about the question, the model must generate the \textit{ideal} answer. %Gold answers are provided and the performance of the model is assest by computing the Rouge score between the generated and gold answers. 
A set of ideal gold answers are provided to assess the performance of the models. We machine translated the questions and ideal answers into French, Italian and Spanish using the NLLB200 3B parameter model \cite{DBLP:journals/corr/abs-2207-04672}.

\section{Experimental Setup}

\begin{figure}
    \centering
    \includegraphics[width=\linewidth]{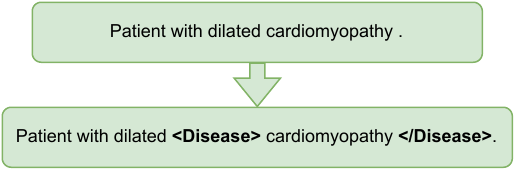}
    \caption{Text-to-Text representation of the Sequence Labeling task. Given an input sentence, the model is expected to generate the same sentence annotated with html-style tags.}
    \label{fig:SL}
\end{figure}

\begin{figure}
    \centering
    \includegraphics[width=\linewidth]{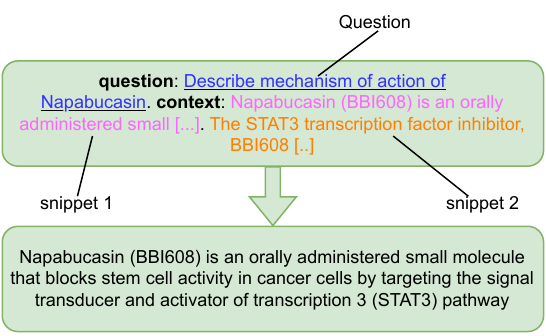}
    \caption{Text-to-Text representation of the BioASQ task. Given a question and a set of relevant snippets, the model generates an answer.}
    \label{fig:BioASQ}
\end{figure}

Medical mT5 is a text-to-text model. This means that, given a text input, it learns to generate a text as output. Therefore, every evaluation task must be converted into a text-to-text format \cite{xue-etal-2021-mt5}. In our experiments the output text is always generated using beam search with 4 beams. 

The list of tasks used for evaluation is listed in Table \ref{tab:tasks}. The \textbf{Sequence Labelling tasks} include medical NER, detecting and classifying named entities according to some pre-defined categories, and Argument Mining, described in Section \ref{sec:new-benchmarks}. Performance for every sequence labelling task is evaluated using standard sequence level $F_1$ score \cite{tjong-kim-sang-de-meulder-2003-introduction}. 

In order to address sequence labelling tasks, text-to-text models such as Medical mT5 are prompted with the sentence to label. As illustrated in Figure \ref{fig:SL}, the expected output is the same sentence annotated with HTML-style tags. The HTML tags for each task are added as special tokens to the model vocabulary. Furthermore, we use constrained decoding to ensure that the output contains the same words as the input and a valid HTML annotation. We use the \textit{Sequence Labeling with LLMs}\footnote{\url{https://github.com/ikergarcia1996/Sequence-Labeling-LLMs}} library. 

With respect to the BioASQ \textbf{Abstractive Question Answering task}, the input prompt contains the question and a context. As shown in Figure \ref{fig:BioASQ}, the context is generated by concatenating all the provided possible snippets. The expected output should be the generated answer to the question, which is then compared to the gold ideal answer.

\begin{table*}[htb]
\centering
\small
\adjustbox{max width=\textwidth}{
\begin{tabular}{@{}llccccccc|cc@{}}
\toprule
Lang & Dataset & mT5\textsubscript{large} & mT5\textsubscript{XL} & SciFive & FlanT5\textsubscript{large} & FlanT5\textsubscript{XL} & mDeBERTa\textsubscript{V3 base}  & BioBERT  & MedMT5\textsubscript{large} & MedMT5\textsubscript{XL} \\ \midrule
EN & NCBI-Disease & 85.1 & 87.7 & \textbf{89.4} & 88.6 & 89.3 & 85.7 & 87.4 & 89.1 & 87.2 \\ \midrule
EN & BC5CDR Disease & 78.5 & 81.4 & 85.4 & 85.0 & \textbf{85.8} & 82.5 & 84.3 & 84.4 & 82.4 \\
EN & BC5CDR Chemical & 89.1 & 90.8 & \textbf{93.3} & 92.0 & 92.9 & 91.1 & 92.9 & 92.8 & 91.3 \\ \midrule
EN & DIANN & 70.1 & 77.8 & 71.9 & 74.4 & 74.2 & \textbf{80.3} & 79.0 & 74.8 & 77.6 \\
\rowcolor{CornflowerBlue!15}ES & DIANN & 72.4 & 74.9 & 70.5 & 70.7 & 70.9 & \textbf{78.3} & 70.2 & 74.9 & 74.8 \\ \midrule
EN & E3C & 54.3 & 60.1 & 62.8 & \textbf{64.2} & 63.1 & 58.2 & 58.6 & 59.4 & 57.9 \\
\rowcolor{CornflowerBlue!15}ES & E3C & 61.6 & 71.7 & 62.7 & 64.4 & 67.1 & 65.9 & 57.4 & \textbf{72.2} & 69.5 \\
\rowcolor{CornflowerBlue!15}FR & E3C & 55.6 & 64.9 & 61.7 & 65.2 & 64.3 & 62.0 & 53.3 & 65.2 & \textbf{65.8} \\
\rowcolor{CornflowerBlue!15}IT & E3C & 61.8 & 63.8 & 59.6 & 61.9 & 65.1 & 63.9 & 52.1 & \textbf{67.5} & 65.9 \\ \midrule
\rowcolor{CornflowerBlue!15}ES & PharmaCoNER & 86.3 & 90.6 & 87.5 & 88.5 & 89.1 & 89.4 & 88.6 & \textbf{90.8} & 90.1 \\ \midrule
EN            & Neoplasm               & 70.4      & 71.1   & 74.4          & \textbf{74.3} & 73.4          & 64.5            & 67.5 & 73.9          & 73.2          \\
EN            & Glaucoma               & 70.7      & 75.1   & 77.1          & \textbf{78.4} & 78.0          & 71.2            & 74.8 & 76.2          & 76.4          \\
EN            & Mixed                  & 68.5      & 73.0   & 73.4          & 73.2          & \textbf{74.5} & 63.4            & 69.6 & 72.2          & 72.0          \\
\rowcolor{CornflowerBlue!15}ES            & Neoplasm               & 69.0      & 56.1   & 71.4          & 72.5          & \textbf{73.9} & 63.0            & 57.1 & 72.1          & 71.8          \\
\rowcolor{CornflowerBlue!15}ES            & Glaucoma               & 69.3      & 70.7   & 73.9          & 73.8          & 75.2          & 68.6            & 64.5 & \textbf{77.1} & 75.5          \\
\rowcolor{CornflowerBlue!15}ES            & Mixed                  & 68.4      & 66.2   & 69.2          & 69.3          & 71.6          & 61.3            & 58.9 & \textbf{72.4} & 71.4          \\
\rowcolor{CornflowerBlue!15}FR            & Neoplasm               & 70.5      & 66.6   & \textbf{74.0} & 72.4          & 73.7          & 63.9            & 59.0 & 72.9          & 71.2          \\
\rowcolor{CornflowerBlue!15}FR            & Glaucoma               & 71.1      & 69.2   & 77.8          & 74.8          & 77.2          & 60.3            & 65.6 & \textbf{79.5} & 75.8          \\
\rowcolor{CornflowerBlue!15}FR            & Mixed                  & 68.3      & 65.4   & 72.0          & 70.9          & \textbf{74.3} & 64.1            & 61.3 & 73.3          & 69.7          \\
\rowcolor{CornflowerBlue!15}IT            & Neoplasm               & 68.1      & 69.9   & 70.1          & 70.9          & 72.0          & 64.4            & 54.8 & 71.2          & \textbf{73.1} \\
\rowcolor{CornflowerBlue!15}IT            & Glaucoma               & 69.2      & 71.5   & 73.7          & 74.0          & 75.9          & 74.7            & 65.8 & 75.7          & \textbf{78.7} \\
\rowcolor{CornflowerBlue!15}IT            & Mixed                  & 66.3      & 67.7   & 67.4          & 69.9          & 70.0          & 61.3            & 57.4 & 70.6          & \textbf{71.9} \\ \midrule
\rowcolor{ForestGreen!10}\multicolumn{2}{c}{AVERAGE}            & 70.2      & 72.1   & 73.6          & 74.1          & 75.1          & 69.9            & 67.3                     & \textbf{75.4} & 74.7          \\
\rowcolor{ForestGreen!10}\multicolumn{2}{l}{AVERAGE ES, FR, IT} & 68.4      & 69.2   & 70.8          & 71.4          & 72.9          & 67.2            & 61.9                     & \textbf{74.0} & 73.2   \\ \bottomrule

\end{tabular}
}
\caption{Single-task supervised F1 scores for Sequence Labelling.}
\label{tab:SingleTask}
\end{table*}

\subsection{Baselines}

As we have developed Medical mT5 by continuing the training of mT5 checkpoints, our primary point of comparison should be mT5 \cite{xue-etal-2021-mt5}. Thus, our first objective
is to assess whether training the model on our multilingual medical-domain
corpus enhances its performance for tasks specific to this domain. Furthermore, we also benchmark our model against SciFive (Pubmed+PMC) a T5-based 738M parameter model
\cite{DBLP:journals/corr/abs-2106-03598} trained exclusively on a corpus of 78B words containing scientific and medical English data. Additionally, we compare the performance of Medical mT5 with Flan-T5 \cite{chung-flan-instruction-models}, which also adopts the T5 architecture but has been finetuned on a huge instruction-following dataset for almost 2K tasks. Flan-T5 achieves state-of-the-art performance in numerous benchmarks, including some from the medical domain \cite{singhal-palm}. We tested all three types of text-to-text models under identical settings and hyperparameters.

We also measure Medical mT5 with the performance of encoder-only models in sequence labelling tasks. We report results with mDeBERTaV3 \cite{debertav3} which is widely used for sequence labeling and excels in multilingual tasks \cite{adelani-etal-2022-masakhaner,Agerri2022LessonsLF}. Although we also tested XLM-RoBERTa \cite{DBLP:conf/acl/ConneauKGCWGGOZ20} and GLOT500 \cite{DBLP:conf/acl/ImaniLKSSKMSMYS23}, their results were worse than those obtained by mDeBERTaV3. Finally, we also compare with BioBERT v1.1 \cite{DBLP:journals/bioinformatics/LeeYKKKSK20}, which has been pretrained on a large English-only biomedical dataset. We do not evaluate the performance of encoder-only models in the question answering task, as their architecture is not designed for text generation.

The specific hyperparameter settings used to fine-tune the models will be available in the Appendix upon publication.

%add info about hyperparameters

\section{Experimental Results}

In this section, we report on the performance of Medical mT5 and of the baselines in the
\textbf{sequence labelling tasks} across different settings. Due to space constraints, we
only report the best performing results.

%The complete results
%including all the models can be found in Appendix \ref{sec:Extended}. 

\paragraph{Single Task Monolingual Supervised Results:} The results when fine-tuning and evaluating the models for each dataset and language are shown in Table \ref{tab:SingleTask}.  The first observation is that Medical-mT5-large significantly outperforms both mT5-large and mT5-XL, demonstrating the benefits of further training these models with our multilingual medical domain corpus.

When comparing Medical mT5 with FlanT5 and SciFive, the latter models are systematically superior on English. This was anticipated since both have been pre-trained with a much larger amount of English-only data specific to the medical domain. With respect to encoder-only models, they achieve in general worse results than text-to-text models across all tasks and languages (except for the DIANN dataset). It is also noteworthy that FlanT5-XL exhibits robust performance across all datasets and languages, even though it was fine-tuned with English-only data not specific to the medical domain. Nonetheless, Medical-mT5-large obtains in general better results for French, Spanish and Italian while being much smaller in size (738M parameters vs 3B parameters), showing the impact of training Medical mT5 with domain-specific data for those languages.

\begin{table*}[htb]
\centering
\small
\adjustbox{max width=0.8\textwidth}{
\begin{tabular}{@{}llccc|ccc@{}}
\toprule
\multirow{2}{*}{Lang} & \multirow{2}{*}{Dataset} & \multicolumn{3}{c}{Single Task} & \multicolumn{3}{c}{MultiTask} \\ 
 &  &  FlanT5\textsubscript{XL} & MedMT5\textsubscript{large} & MedMT5\textsubscript{XL} &  FlanT5\textsubscript{XL} & MedMT5\textsubscript{large} & MedMT5\textsubscript{XL} \\ \midrule
EN & NCBI-Disease & \textbf{89.3} & 89.1 & 87.2 & 87.6 & 87.6 & 86.9 \\ \midrule
EN & BC5CDR Disease & \textbf{85.8} & 84.4 & 82.4 & 85.1 & 83.4 & 83.0 \\
EN & BC5CDR Chemical & \textbf{92.9} & 92.8 & 91.3 & 92.7 & 92.5 & 91.6 \\ \midrule
EN & DIANN & 74.2 & 74.8 & 77.6 & \textbf{80.0} & 75.4 & 75.3 \\
\rowcolor{CornflowerBlue!15}ES & DIANN & 70.9 & 74.9 & 74.8 & \textbf{77.1} & 72.6 & 73.6 \\ \midrule
EN & E3C & \textbf{63.1} & 59.4 & 57.9 & 62.1 & 60.9 & 62.0 \\
\rowcolor{CornflowerBlue!15}ES & E3C & 67.1 & 72.2 & 69.5 & 66.5 & \textbf{74.9} & 73.3 \\
\rowcolor{CornflowerBlue!15}FR & E3C & 64.3 & 65.2 & \textbf{65.8} & 62.9 & 65.4 & 65.1 \\
\rowcolor{CornflowerBlue!15}IT & E3C & 65.1 & \textbf{67.5} & 65.9 & 60.7 & 66.9 & 65.1 \\ \midrule
\rowcolor{CornflowerBlue!15}ES & PharmaCoNER & 89.1 & \textbf{90.8} & 90.1 & 89.9 & 90.3 & 89.5 \\ \midrule
EN & Neoplasm & 73.4 & \textbf{73.9} & 73.2 & 73.1 & 72.3 & 72.9 \\
EN & Glaucoma & \textbf{78.0} & 76.2 & 76.4 & 76.4 & 76.8 & 77.5 \\
EN & Mixed & \textbf{74.5} & 72.2 & 72.0 & 71.5 & 70.9 & 73.0 \\
\rowcolor{CornflowerBlue!15}ES & Neoplasm & \textbf{73.9} & 72.1 & 71.8 & 73.5 & 73.5 & 73.7 \\
\rowcolor{CornflowerBlue!15}ES & Glaucoma & 75.2 & 77.1 & 75.5 & 77.1 & 77.7 & \textbf{79.3} \\
\rowcolor{CornflowerBlue!15}ES & Mixed & 71.6 & 72.4 & 71.4 & 70.0 & 71.8 & \textbf{72.8} \\
\rowcolor{CornflowerBlue!15}FR & Neoplasm & 73.7 & 72.9 & 71.2 & \textbf{74.0} & 72.9 & 73.6 \\
\rowcolor{CornflowerBlue!15}FR & Glaucoma & 77.2 & \textbf{79.5} & 75.8 & 76.6 & 77.0 & 79.4 \\
\rowcolor{CornflowerBlue!15}FR & Mixed & \textbf{74.3} & 73.3 & 69.7 & 71.8 & 71.2 & 73.0 \\
\rowcolor{CornflowerBlue!15}IT & Neoplasm & 72.0 & 71.2 & 73.1 & 71.9 & \textbf{74.6} & 74.0 \\
\rowcolor{CornflowerBlue!15}IT & Glaucoma & 75.9 & 75.7 & 78.7 & 77.6 & 78.5 & \textbf{78.9} \\
\rowcolor{CornflowerBlue!15}IT & Mixed & 70.0 & 70.6 & 71.9 & 69.9 & 72.5 & \textbf{73.3} \\ \midrule
\rowcolor{ForestGreen!10}\multicolumn{2}{c}{AVERAGE} & 75.1 & 75.4 & 74.7 & 75.2 & 76.2 & \textbf{76.7} \\
\rowcolor{ForestGreen!10}\multicolumn{2}{c}{AVERAGE ES, FR, IT} & 72.9 & 74.0 & 73.2 & 73.1 & 74.8 & \textbf{75.3} \\ \bottomrule
\end{tabular}
}
\caption{Multi-task supervised F1 scores for Sequence Labelling.}
\label{tab:MultiTask}
\end{table*}

\paragraph{Multi-Task Supervised Results:} Text-to-text models have
demonstrated improved performance when trained in multi-task settings
\cite{chung-flan-instruction-models}. Following this, we also experimented with fine-tuning them across all the sequence labeling tasks simultaneously. To inform the model about which labels should classify for each input example, we add the list of predefined labels from the corresponding dataset to the beginning of the input sentence. For instance, the input depicted in Figure \ref{fig:SL} is adjusted to \textit{``<Disease> Patient with dilated cardiomyopathy''}. A comparison of the Single Task and Multi-Task settings is presented in Table \ref{tab:MultiTask}. It can be seen that in this setting Medical mT5 achieves the best overall results for Spanish, French and Italian. On average, Medical-mT5-xl also obtains the best performance, slightly improving over the results of FlanT5-XL and Medical-mT5-large.

\begin{table*}[htb]
\centering
\small
\adjustbox{max width=0.8\linewidth}{
\begin{tabular}{@{}llcccc|cc@{}}
\toprule
Lang & Dataset & mT5\textsubscript{XL} & SciFive & FlanT5\textsubscript{XL} & mDeBERTa\textsubscript{V3 base} & MedMT5\textsubscript{large} & MedMT5\textsubscript{XL} \\ \midrule
ES         & Neoplasm       & 71.4          & 69.8    & 67.9      & 65.1            & \textbf{72.4} & 71.7          \\
ES         & Glaucoma       & \textbf{74.1} & 71.5    & 70.6      & 68.3            & 72.4          & 73.2          \\
ES         & Mixed          & \textbf{69.4} & 67.0    & 66.7      & 60.9            & 68.1          & 68.8          \\
FR         & Neoplasm       & 71.6          & 68.6    & 69.9      & 60.5            & 72.4          & \textbf{72.8} \\
FR         & Glaucoma       & 75.8          & 74.5    & 71.0      & 68.7            & 72.3          & \textbf{76.7} \\
FR         & Mixed          & \textbf{73.0} & 68.5    & 68.2      & 59.3            & 70.4          & 72.4          \\
IT         & Neoplasm       & 70.6          & 63.1    & 67.3      & 62.4            & 72.9          & \textbf{73.2} \\
IT         & Glaucoma       & 76.7          & 71.6    & 72.0      & 70.2            & 75.4          & \textbf{79.0} \\
IT         & Mixed          & 69.9          & 62.5    & 66.9      & 62.1            & 71.7          & \textbf{71.9} \\ \midrule
\rowcolor{ForestGreen!10} \multicolumn{2}{c}{AVERAGE} & 72.5          & 68.6    & 69.0      & 64.2            & 72.0          & \textbf{73.3} \\ \bottomrule
\end{tabular}
}
\caption{Zero-shot F1 scores for Argument Mining. Models have been trained in English and evaluated in Spanish, French and Italian.}
\label{tab:ZeroShot}
\end{table*}

\paragraph{Zero-shot Cross-Lingual Transfer Results:} Manually annotated medical domain datasets for languages other than English are scarce. Therefore, developing models that can successfully generate predictions for languages different to those used for fine-tuning is crucial. We evaluate this ability to perform zero-shot cross-lingual transfer by fine-tuning Medical mT5 and the baselines on the English AbsRCT Neoplasm dataset, and then evaluating them on the Neoplasm, Glaucoma, and Mixed datasets for Spanish, French, and Italian. The results are presented in Table \ref{tab:ZeroShot}. Results show that Medical mT5 outperforms any other model. Moreover, Medical-mT5-xl achieves significantly better results than Medical-mT5-large. 

To summarize, Medical mT5 stands out for its superior performance in the evaluation for Spanish, French, and Italian languages, especially for the multitask and the zero-shot transfer settings. These capabilities can help mitigate the scarcity of manually annotated medical data for other target languages. In contrast, SciFive and FlanT5, having been trained on extensive English-only datasets, emerge as the top choices when the primary focus is on English-only tasks.

Finally, despite Medical-mT5-xl being larger than Medical-mT5-large (3B vs 738M), its performance is worse in the single-task evaluation setting. This behaviour is not observed in the multi-task and zero-shot experiments, leading us to hypothesize that the larger Medical-mT5-xl model is more prone to overfit in the single-task supervised setting.

\begin{table*}[htb]
\centering
\adjustbox{max width=0.9\textwidth}{

\begin{tabular}{@{}lp{16cm}@{}}
\toprule
\rowcolor{CornflowerBlue!15} Ideal Answer & Which type of urinary incontinence is diagnosed with the Q tip test?  Stress urinary incontinence is diagnosed with the Q tip test. The test evaluates urethral mobility.  \\ \midrule
mT5-large & Q-tip test is used for urodynamic assessment of patients with severe urinary incontinence.  \\
FlanT5-large & The Q tip test is used for assessing urethral mobility.  \\
SciFive & Q tip test is used for evaluation of stress urinary incontinence. \\
MedMT5-large & Q-tip test is used for evaluation of stress urinary incontinence.  \\ \midrule
\rowcolor{CornflowerBlue!15} Ideal Answer & Which are the main manifestations of Ohdo syndrome? Severe ID, absent or deficient language, skeletal manifestations including bilateral patella dislocations.  \\ \midrule
mT5-large & Skeletal manifestations in Ohdo syndrome are a case with bilateral patella dislocations where surgical intervention has been indicated.  \\
FlanT5-large & The main manifestations of Ohdo syndrome are: 1) severe ID, 2) absent or deficient language and 3) milder, clinical manifestation in heterozygotes. \\
SciFive & Ohdo syndrome is characterized by severe ID, absent or deficient language and, milder, clinical manifestation in heterozygotes.  \\
MedMT5-large & The main manifestations of Ohdo syndrome are: 1) absent or deficient language and 2) mildder clinical manifestation in heterozygotes.  \\ \bottomrule
\end{tabular}}
\caption{Examples of answers generated by each model for two different BioASQ questions together with the rank assigned by medics.}
\label{tab:BioASQ_example}
\end{table*}

\subsection{Abstractive Question Answering}

In this section, we explore the text generation capabilities of Medical mT5 and other baseline text-to-text models on the BioASQ question answering dataset described in Section
\ref{sec:QA_explained}. Previous work typically evaluate the performance 
on this task using the ROUGE score \cite{bioasq} to
compare the gold standard answer with the answer generated by the model.
However, we find this metric inadequate for medical domain tasks as it does not
address crucial aspects of the generation such as factuality, potential harm, and bias
\cite{singhal-palm}. Consequently, we enlisted medical
professionals to analyze the answers produced by the models.

During annotation, medical doctors were displayed the question, the ideal gold answers and
the answers generated by each model. If required, they could also inspect
the snippets that provide context to answer each of the questions. We narrowed the
evaluation to Medical-mT5-large, mT5-large, FlanT5-large and SciFive. The evaluation
was conducted by medical doctors proficient/native speakers of English, French and Spanish. For each question, doctors were asked to rank the
answers generated by the models as the best, second-best, third-best, and worst
answer.

Two Spanish medical doctors proficient or native in English and Spanish analyzed 50 English examples and 252 Spanish. For the French language, 3 French clinicians analyzed 186 answers, of which 47 were done by 2 doctors to calculate IAA (Cohen's Kappa Score: 0.28 and Average Spearman's Rank Correlation: 0.48), which indicates a low level of agreement. This exercise provided interesting insights with respect to the performance of the models in text generation tasks in the medical domain. First, medical doctors could not in general establish significant differences between the
answers generated by each of the models; predictions were far too similar, and all tended to
fail on the same questions. As an example, Table \ref{tab:BioASQ_example} shows the answers to two different questions. As it can be observed, the answers generated by each model are very similar, and the doctors ended up ranking them primarily based on style. 

%This is reflected in the number of "wins" for each system. In English, FlanT5-large was selected as the best model 14 times, SciFive 14 times, MT5-large 10 times, and MedicalMT5-large 7 times. \textcolor{red}{For French, FlanT5-large was selected as the best model 75 times, SciFive 63 times, MedicalMT5-large 59 times, and MT5-large 55 times.} For Spanish, MT5-large was selected as the best model 47 times, Medical mT5 36 times, SciFive 29 times, and FlanT5 19 times.

The final result of the manual analysis is that all the models were chosen a similar number of times as the best. 
%However, it is noteworthy that FlanT5 and SciFive, which have tokenizers trained only on English data, skip accents in French or Spanish, resulting in words missing letters. 
We believe that this demonstrates the difficulty of performing and obtaining meaningful evaluation results for this kind of tasks on this specific domain. This is in fact supported by the low IAA agreement obtained in the French annotation. This issue has also emerged in prior research and was partially addressed by employing a very large number of experts and asking them to respond with a yes/no to a set of predefined potential issues in the model output \cite{singhal-palm}. Still, the variance on the answers provided by the experts was significant.

However, there could be other underlying reasons for this behaviour. First, perhaps the T5 architecture is not ideally suited for text generation as formulated in the BioASQ task, as these models are trained on a masking reconstruction objective rather than on direct text generation tasks. Consequently, the knowledge acquired during pre-training might not generalize well when the models are subsequently trained for text generation purposes. Second, perhaps using much larger models such as MedPaLM \cite{singhal-palm} may generate better answer generation, but models of 540B parameters are currently unusable for the large majority of the NLP research labs, including ours. Nonetheless, it should be stressed that research on appropriate evaluation metrics for these tasks is still a difficult challenge which requires further investigation. 

In any case, our results demonstrate the potential of a text-to-text model such as Medical mT5 for multilingual sequence labelling in the medical domain, establishing new state-of-the-art results in the multi-task and zero-shot cross-lingual settings.

% This will facilitate future training and evaluation of models geared towards text generation (such as Causal Language Models) . 

\section{Conclusion}

In this paper we have presented Medical mT5, the first open source multilingual text-to-text LLM for the medical domain. Its development has required the compilation of a new 3B word corpus in English, French, Italian and Spanish specific to the medical domain. Furthermore, motivated by the lack of multilingual benchmarks, we have generated evaluation benchmarks for French, Italian and Spanish for Argument Mining and Abstractive Question Answering. 

With respect to the languages chosen in this paper, we would like to comment that acquiring medical domain data is extremely challenging, even for languages such as the ones included. Furthermore, the choice of languages was also influenced by the availability of native medical doctors to do the manual evaluation for Abstractive Question Answering. In any case, we hope that our paper will encourage more researchers to join our effort and gather data for their respective languages, thereby creating larger, multilingual medical domain datasets encompassing more languages in the future. 

A comprehensive experimentation on sequence labelling tasks shows that Medical mT5 outperforms strong text-to-text baselines of similarly-sized models in the multi-task and zero-shot cross-lingual evaluation settings. This is particularly interesting as these settings fully exploit the multilingual nature of a text-to-text model such as Medical mT5.

Furthermore, our experiments on Abstractive Question Answering show the inherent difficulty of evaluating generative tasks for this specific domain, where complex issues such as truthfulness and veracity are difficult to capture by automatic metrics. Manual evaluation is not ideal either, as medical doctors were not able to clearly distinguish between the quality of the answers generated by the different models. In line with previous work \cite{singhal-palm}, we hope our paper will bring further attention to this problem and encourage further research on evaluation methods.

\section{Acknowledgements}

\noindent \textbf{HiTZ Center}: This work has been supported by the following MCIN/AEI/10.13039/501100011033 projects: (i) Antidote (PCI2020-120717-2) and EU NextGenerationEU/PRTR (ii) DeepKnowledge (PID2021-127777OB-C21) and by FEDER, EU; (iii) DeepR3 (TED2021-130295B-C31) and EU NextGeneration EU/PRTR. Iker García-Ferrero is supported by a doctoral grant from the Basque Government (PRE\_2021\_2\_0219). Rodrigo Agerri currently holds the RYC-2017-23647 fellowship (MCIN/AEI/10.13039/501100011033 and by ESF Investing in your future).

\noindent \textbf{FBK}: This work has been supported by the European Union under Horizon Europe Projects ANTIDOTE (PCI2020-120717-2), eCREAM (Grant No. 101057726) and IDEA4RC (Grant No. 101057048).
Views and opinions expressed are however those of the author(s) only and do not necessarily reflect those of the European Union.

\noindent \textbf{UCA}: This work has been supported by the French government, through the 3IA Côte d’Azur Investments in the Future project managed by the National Research Agency (ANR) with the reference number ANR-19P3IA-0002. This work was supported by the CHISTERA grant of the Call XAI 2019 of the ANR with the grant number Project-ANR-21-CHR4-0002.

\section{Ethical Statement}

Our research in developing Medical mT5, a multilingual text-to-text model for the medical domain, has ethical implications that we acknowledge. Firstly, the broader impact of this work lies in its potential to improve medical communication and understanding across languages, which can enhance healthcare access and quality for diverse linguistic communities. However, it also raises ethical considerations related to privacy and data security. To create our multilingual corpus, we have taken measures to anonymize and protect sensitive patient information, adhering to data protection regulations in each language's jurisdiction or deriving our data from sources that explicitly address this issue in line with privacy and safety regulations and guidelines. Furthermore, we are committed to transparency and fairness in our model's development and evaluation. We have worked to ensure that our benchmarks are representative and unbiased, and we will continue to monitor and address any potential biases in the future. Finally, we emphasize our commitment to open source by making our data, code, and models publicly available, with the aim of promoting collaboration within the research community.

\nocite{*}
\section*{Bibliographical References}\label{sec:reference}

\bibliographystyle{lrec-coling2024-natbib}
\bibliography{references}

\section*{Language Resource References}
\label{lr:ref}
\bibliographystylelanguageresource{lrec-coling2024-natbib}
\bibliographylanguageresource{languageresource}

%\appendix
%%% APPENDIX IS NOT ALLOWED DURING REVIEW. 

%\section{Extended results}
%\label{sec:Extended}

%\input{tables/singletask_extended}
%\input{tables/multitask_extended}
%\input{tables/zeroshot_extended}

\end{document}